\def\year{2022}\relax
\documentclass[letterpaper]{article} % DO NOT CHANGE THIS
\usepackage{aaai22}  % DO NOT CHANGE THIS
\usepackage{times}  % DO NOT CHANGE THIS
\usepackage{helvet}  % DO NOT CHANGE THIS
\usepackage{courier}  % DO NOT CHANGE THIS
\usepackage[hyphens]{url}  % DO NOT CHANGE THIS
\usepackage{graphicx} % DO NOT CHANGE THIS
\usepackage{natbib}  % DO NOT CHANGE THIS AND DO NOT ADD ANY OPTIONS TO IT
\usepackage{caption} % DO NOT CHANGE THIS AND DO NOT ADD ANY OPTIONS TO IT
\usepackage{multirow}
\usepackage{adjustbox}
\DeclareCaptionStyle{ruled}{labelfont=normalfont,labelsep=colon,strut=off} % DO NOT CHANGE THIS
\usepackage{algorithm}
\usepackage{algorithmic}
\usepackage{newfloat}
\usepackage{listings}
\usepackage{amsfonts}

\floatstyle{ruled}
\newfloat{listing}{tb}{lst}{}
\floatname{listing}{Listing}
\title{Prior Omission of Dissimilar Source Domain(s) \\ for Cost-Effective Few-Shot Learning}
\author{ Zezhong Wang\footnotemark[1], 
        Hongru Wang\footnotemark[1], 
        Kwan Wai Chung,\\
        Jia Zhu,
        Gabriel Pui Cheong Fung, 
        Kam-Fai Wong \\
    Department of Systems Engineering and Engineering Management \\
    The Chinese University of Hong Kong \\
   \{zzwang, kfwong\}@se.cuhk.edu.hk \\
}

\usepackage{bibentry}
\usepackage{xcolor}
\usepackage{amsmath}
\definecolor{mypink1}{rgb}{0.858, 0.188, 0.478}
\begin{document}

\maketitle
\footnotetext[1]{Equal contributions.}
\begin{abstract}
Few-shot slot tagging is an emerging research topic in the field of Natural Language Understanding (NLU). With sufficient annotated data from source domains, the key challenge is how to train and adapt the model to another target domain which only has few labels. Conventional few-shot approaches use all the data from the source domains without considering inter-domain relations and implicitly assume each sample in the domain contributes equally. However, our experiments show that the data distribution bias among different domains will significantly affect the adaption performance. Moreover, transferring knowledge from dissimilar domains will even introduce some extra noises so that affect the performance of models. To tackle this problem, we propose an effective similarity-based method to select data from the source domains. In addition, we propose a Shared-Private Network (SP-Net) for the few-shot slot tagging task. The words from the same class would have some shared features. We extract those shared features from the limited annotated data on the target domain and merge them together as the label embedding to help us predict other unlabelled data on the target domain. The experiment shows that our method outperforms the state-of-the-art approaches with fewer source data. The result also proves that some training data from dissimilar sources are redundant and even negative for the adaption.
\end{abstract}
\section{Introduction}

\begin{figure}[t]
  \centering
  \includegraphics[scale=0.32]{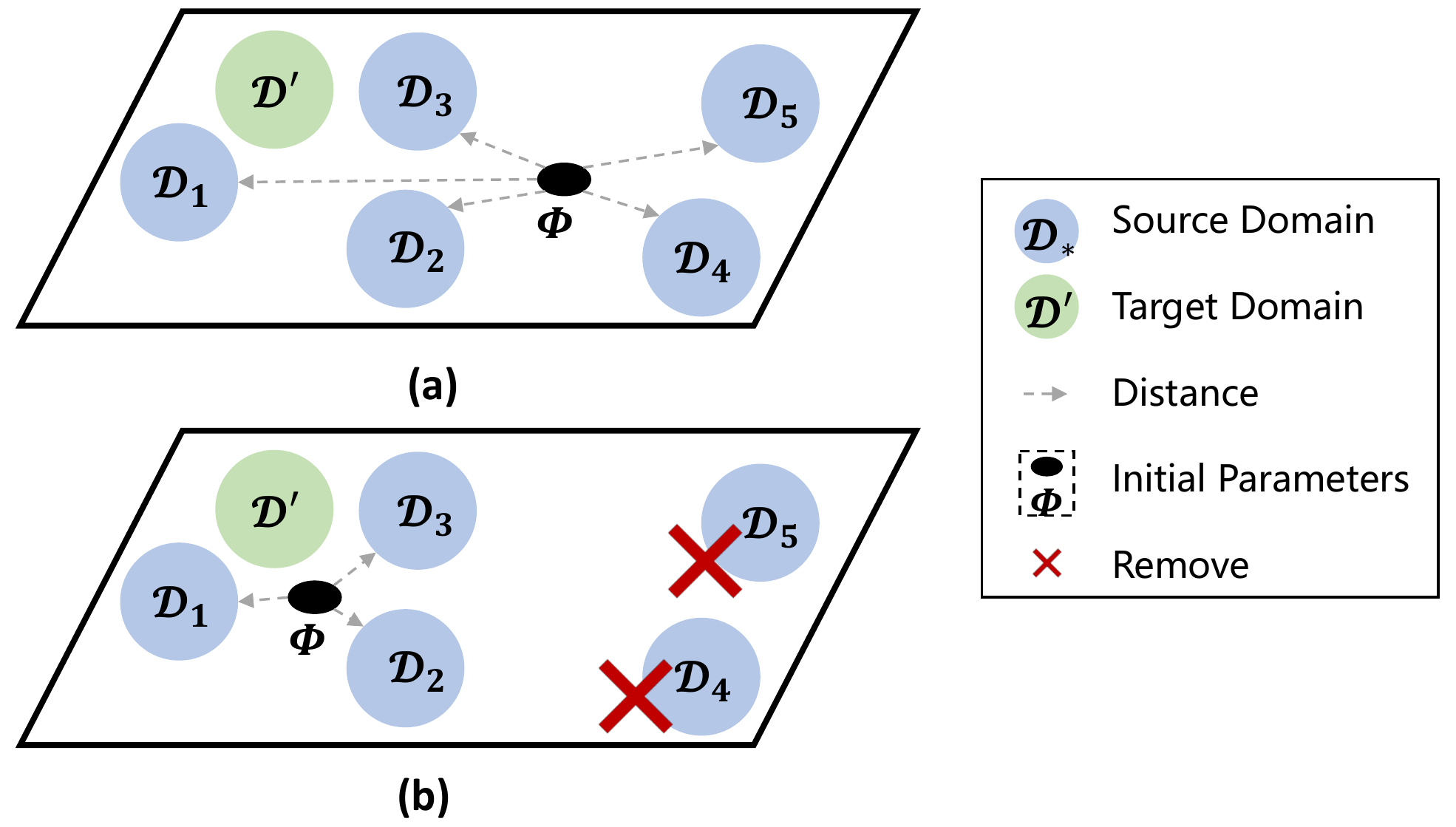}
  \caption{The difference between training with (a) all data and (b) data selection. The dashed line represents the distance among different domains in the parameter space with the centroid ($\Phi$). With data selection, we remove the dissimilar domains $\mathcal{D}_4$ and $\mathcal{D}_5$ from training and the centroid will be closer to the target domain $\mathcal{D}^{\prime}$.}
  \label{data_selection}
\end{figure}

Slot tagging \cite{tur2011spoken}, one of the crucial problems in Natural Language Understanding (NLU), aims to recognize pre-defined semantic slots from sentences and usually is regarded as a sequence labeling problem \cite{sarikaya2016overview}. For example, given a sentence ``Book a ticket to London", the word ``London" should be recognized as the slot ``CITY" by NLU model.

 Currently, most of the methods for the slot tagging task have a notorious limitation that they requires a lot of annotated data. However, there are almost infinite long tail domains in the real scenarios \cite{6909517} so that it is nearly impossible to annotate sufficient data for each domain. Therefore, few-shot learning methods \cite{ravi2016optimization} have received attention as it can transfer the knowledge learned from the existing domains to new domains quickly with limited data.
 
Current works \cite{pmlr-v97-yoon19a, liu2020coach, wang2021mcml} proposed various methods to improve the performance of slot tagging few-shot learning, but most of them focus on ``how" to transfer rather than ``what" should be transferred. The knowledge from the not-relevant source domain is hard to help the model identify the slots in the new domain. Further, such kind of knowledge is redundant and sometimes could be regarded as noises that even deteriorates the performance \cite{wang2019characterizing}. We observe this phenomenon and prove the existence of the negative transfer in the experiment. To this end, we propose a similarity-based method to evaluate the inter-domain relation and indicate which domains should be selected for training. Specifically, we calculate three different similarities including target vocabulary covered (TVC), TF-IDF similarity (TIS), and label overlap (LO) between domains and combine them with different weights. The combined similarity function selects data from both corpus level and label level, which is more comprehensive. In this way, the dissimilar sources will be rejected and the initial parameters of the model will be naturally more closed to the local optimum of the target domain. A high-level intuition of the difference between training with all data and training with data selection is shown in Figure~\ref{data_selection}. 

After selecting proper data, we also propose a solution about ``how" to transfer knowledge for few-shot slot tagging task. Specifically, we build a Shared-Private Network to capture stable label representations under the few-shot setting. Many works \cite{hou2020fewshot, zhu2020vector, liu2020coach} try to enhance the accuracy of slot identification from the label representation engineering. They assign each label with a semantic vector \cite{snell2017prototypical, hou2020fewshot, zhu2020vector, pmlr-v97-yoon19a} rather than a simple one-hot encoding. However, the quality of the label representations highly depends on the volume of the training samples and suffers from the unstable problem under the few-shot setting due to the extremely biased data distribution. Hence, we propose the Shared-Private Network to separate the shared features and private features from the limited samples. The words with the same label share common information. They are extracted and saved as shared features. Other parts are regarded as detailed information related to the words and will be saved as private features. After filtering the detailed information out, the label representation generated according to the shared features will be more robust against the annotation shortage problems in the few-shot setting.

The contributions of this work are as follows:
\begin{itemize}
\item We propose a similarity-based method to measure the relation among domains to guide data selection and to avoid negative knowledge transfer in few-shot learning.
\item We propose the Shared-Private Network to extract more stable label representation with limited annotations.
\item We prove the existence of negative transfer via experiments and give explanations about this phenomenon via visualization.
\end{itemize} 

\section{Related Work}
\label{sec_rw}
Convention studies in slot tagging mainly focus on proposing and utilizing deep neural networks to recognize the semantic slots in given contexts \cite{shi2016recurrent, kim2017speaker}. However, most of these models need a large amount of annotated data which is quite scarce in the real world, especially for those minority domains. Recent works \cite{bapna2017zeroshot,shah2019robust,rastogi2019towards, liu2020coach} propose several few-shot learning methods for slot tagging and developed domain-speciﬁc model with limited annotated data. \citet{hou2020fewshot} introduced a collapsed dependency transfer mechanism into the conditional random ﬁeld (CRF) and proposed the Label-enhanced Task-Adaptive Projection Network (L-TapNet) which build a strong few-shot baseline for slot tagging. Based on the work of \citet{hou2020fewshot}, \citet{zhu2020vector} then introduced a vector projection network for few-shot slot tagging. It is worth to note that, due to the lack of annotation on the target domain, both approaches paid attention to label representation engineering rather than using conventional one-hot encoding directly. But building label representation with limited annotations is still a challenge. To stabilize the effectiveness of label representation, we proposed a Shared-Private network to learn representation from shared information of words.

Besides that, negative transfer that transferring knowledge from the source can have a negative impact on the target has been founded in many tasks \cite{wang2019characterizing, chen2019catastrophic, gui2018negative}. Because of this phenomenon, methods for relation analysis between source and target domains has been proposed recently. \citet{gururangan2020dont} use vocabulary overlap as the similarity between two datasets and emphasized the significant impact of domain-adaptive for pre-training. \citet{dai-etal-2019-using} study different similarity methods including target vocabulary covered (TVC), language model perplexity (PPL), and word vector variance (WVV) to select data for pre-training tasks. However, a single similarity function does not work well in the few-shot setting. Different similarity methods always give diverse data selection strategies and are hardly consistent. To this end, we propose a comprehensive indicator that combines three similarity functions to guide the data selection in the few-shot setting.

\section{Problem Definition}
\label{sec_pd}
We follow the same task definition as \citet{hou2020fewshot}. Given a sentence $\textbf{x} = (x_1, x_2, \cdots, x_n)$ as a sequence of words, slot tagging task aims to assign the corresponding label series $\textbf{y} = (y_1, y_2, \cdots, y_n)$ to indicate which classes the words should belong to. A domain $\mathcal{D}=\{(\textbf{x}^{(i)},\textbf{y}^{(i)})\}_{i=1}^{N_\mathcal{D}}$ is a set of $(\textbf{x}, \textbf{y})$ pairs that from same scenario and $N_\mathcal{D}$ is the number of sentences in domain $\mathcal{D}$.

In few-shot setting, models are trained from source domain $\{\mathcal{D}_1, \mathcal{D}_2, \cdots \}$ and are applied to the target domain $\{\mathcal{D}^{\prime}_1, \mathcal{D}^{\prime}_2, \cdots \}$ which are new to the models. It is worth note that there are only few labeled samples, which make up the support set $\mathcal{S}=\{(\textbf{x}^{(i)},\textbf{y}^{(i)})\}_{i=1}^{N_S}$, in each target domain $\mathcal{D}^{\prime}_j$. For each unique $N$ labels (N-way) in support set $\mathcal{S}$, there are $K$ annotated samples (K-shot). Besides that, the samples in the target domain ${\mathcal{D}^{\prime}_j}$ are unlabeled.

Thus, few-shot slot tagging task is defined as follows: given a K-shot support set $\mathcal{S}$ and a query sentence $\textbf{x} = (x_1, x_2, \cdots, x_n)$, determine the corresponding labels sequence $\textbf{y}^*$:
\begin{equation}
\textbf{y}^* = (y_1^*, y_2^*, \cdots, y_n^*) = \mathop{\arg\max}_{\textbf{y}}p(\textbf{y\big|\textbf{x},$\mathcal{S}$})
\end{equation}
\section{Data Selection}
\label{ds}
In this section, we first show the existence of negative knowledge transfer among domains. The phenomenon demonstrates the necessity of data selection. Then introduce our similarity-based data selection strategy that can be used to avoid negative knowledge transfer to improve performance in few-shot slot tagging.

\subsection{Negative Knowledge Transfer}
\begin{figure}[t]
  \centering
  \includegraphics[scale=0.3]{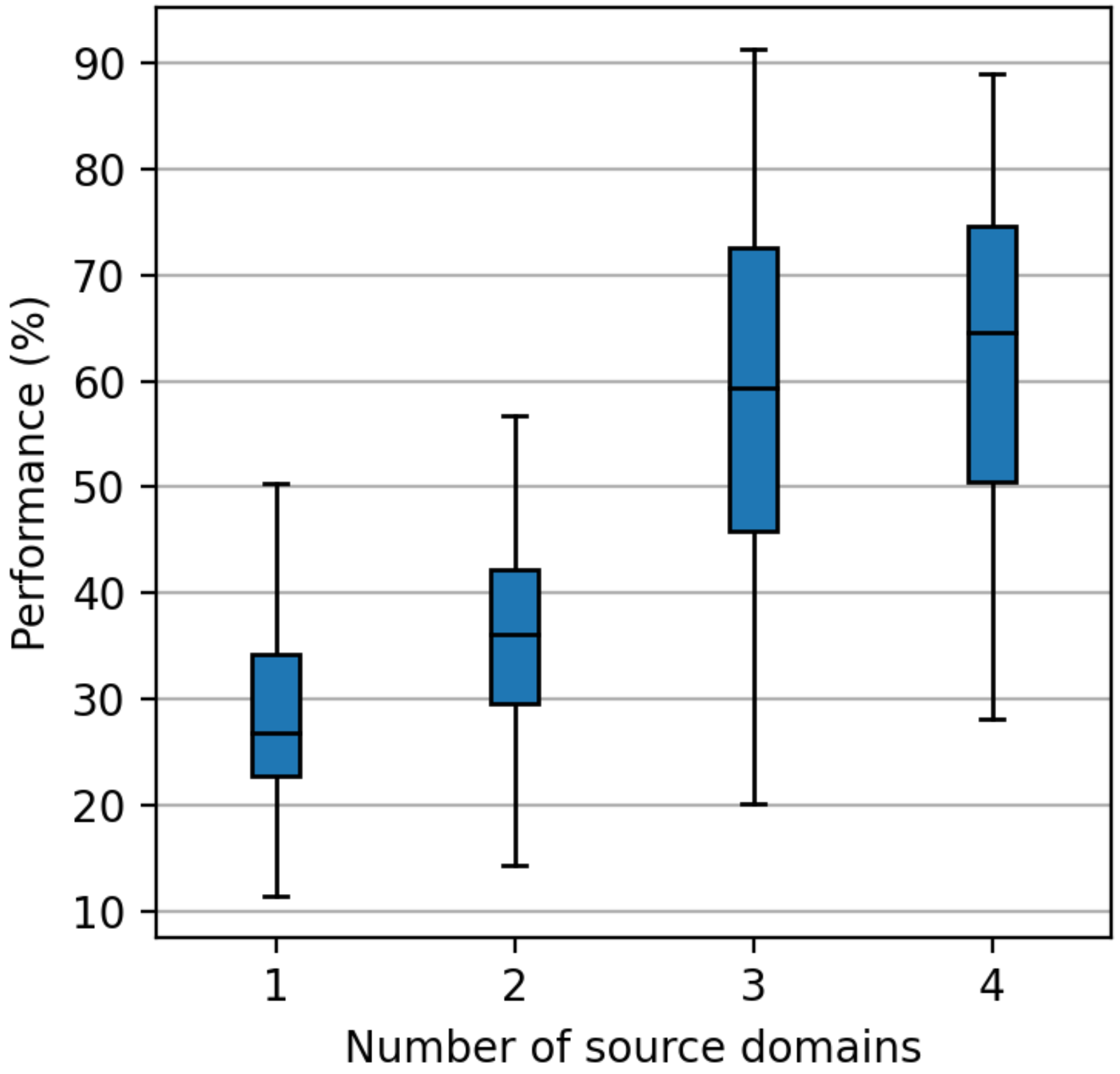}
  \caption{The relationship between performance (y-axis), specifically the F1 score, and the number of source domains (x-axis).}
  \label{boxplot}
\end{figure}

Due to negative knowledge transfer, some knowledge the model learned before is useless and may affect the judgment of the model on the new domains, which will degrade the performance. In the preliminary study, we train the model with all different combinations of source domains and record their performance. The relation between the number of source domains and their corresponding performance is shown in Figure \ref{boxplot}. Overall, with more training domains, the performance would be better. However, comparing the maximum values, it is obvious that training with 3 source domains outperforms training with 4. This phenomenon indicates that more source domains may even decrease the performance and proves the existence of negative knowledge transfer. It also inspires us that the model will achieve a better result with proper data selection.

\subsection{Selection Strategy}

To avoid negative knowledge transfer, an indicator is needed to select data or source domains before training. Given a group of data from source domain and the data of target domain, the indicator should output a score which can reflect how fit are these source data for transferring knowledge to the target. Ideally, the indicator score behaves linearly with the performance so that higher indicator score can lead to better performance. In this way, the group of source data with highest indicator score can be selected as the best choice for training.

% \begin{figure}[h]
%   \centering
%   \includegraphics[scale=0.42]{pics/indicator.pdf}
%   \caption{This diagram shows a good indicator (green line) and a bad indicator (orange line). The green line demonstrate a monotonic increasing linear relationship between the indicator score and performance. On the contrary, the orange line demonstrate a nearly random relationship.}
%   \label{indicator}
% \end{figure}

The data that can be leveraged includes the source domains $\{\mathcal{D}_1, \cdots , \mathcal{D}_M\}$ with sufficient labels, the support set $\mathcal{S}_j$ with labels in the target domain $\mathcal{D}^{\prime}_j$, and the query set $\mathcal{Q}_j$ without labels. Notice that the data in the support set $\mathcal{S}_j$ is much less than the query set $\mathcal{Q}_j$. Considering the attributes mentioned above and the data we can use, we investigate three similarity functions as indicators for data selection.

\textbf{Target Vocabulary Covered (TVC) } is a significant corpus level feature that represents the overlap of vocabulary between source domain(s) and a target domain and is defined as:
\begin{equation}
    \textbf{TVC}(\mathcal{D}_i, \mathcal{D}^{\prime}_j) = \frac{\left|V_{\mathcal{D}_i} \cap V_{\mathcal{D}^{\prime}_j} \right|}{\left|V_{\mathcal{D}^{\prime}_j}\right|}
\end{equation}
where $V_{\mathcal{D}_i}$ and $V_{\mathcal{D}^{\prime}_j}$ are the vocabularies (sets of unique tokens) of the source domain $\mathcal{D}_i$ and the target domain $\mathcal{D}^{\prime}_j$ respectively and $\mathbf{|\cdot|}$ is the norm operation that indicates the size of the set. Intuitively, if most of words in the target domain have already appeared in the sources, the word embeddings should have been well trained so that improves the performance.

\textbf{TF-IDF Similarity (TIS)} is another corpus level feature \cite{bao2020fewshot}. We treat each domain as a document and calculate their \verb|tf-idf| features \cite{salton1988term, wu2008interpreting}. Cosine similarity is used to evaluate the correlation between the sources and the target. Compared with TVC, TIS assign each word with a weight according to the term frequency and inverse document frequency, which takes fine-grained corpus feature into account. The details are shown below:
\begin{equation}
    \text{tf}_{i,j}= \frac{n_{ij}}{\sum_{k}n_{k,j}}
\end{equation}
where $n_{ij}$ is the times of word $t_{i}$ appeared in domain $\mathcal{D}_{j}$.

\begin{equation}
    \text{idf}_{i}= \lg \left(\frac{M}{|\{j:t_{i}\in \mathcal{D}_{j}\}_{j=1}^{M}|}\right)
\end{equation}
where $M$ is the total number of domains. And the \verb|tf-idf| feature is the product of \verb|tf| and \verb|idf|: 
\begin{equation}
\text{tf-idf}_{j} = \text{tf}_{i,j} \cdot \text{idf}_{i}
\end{equation}
$\text{tf-idf}_{j}$ can be regarded as the word distribution feature of the domain $j$ and cosine similarity is used to evaluate the correlation between two domains:

\begin{equation}
\text{TIS}(\mathcal{D}_i, \mathcal{D}_j) = \frac{\text{tfidf}_{\mathcal{D}_i} \cdot \text{tfidf}_{\mathcal{D}_j}}{||\text{tfidf}_{\mathcal{D}_i}||_{2} \cdot ||\text{tfidf}_{\mathcal{D}_j}||_{2}}
\end{equation}
where $|| \cdot ||_{2}$ is the Euclidean norm. 

\textbf{Label Overlap (LO)} is a label level feature that represents the overlap of labels between source domains and the target domain. Although labels are quite scarce in the target domain under the few-shot setting, the types of labels are not. Every label on the target domain at least appeared $K$ times (K-shot) in the support set $\mathcal{S}$ and therefore the types of the labels are complete. Hence, label overlap is also a good choice as data selection indicator:
\begin{equation}
    \text{LO}(Y_i, Y_j) = \frac{\left|Y_i \cap Y_j \right|}{\left|Y_j\right|}
\end{equation}
where $Y_i$ and $Y_j$ stand for the unique label set of the source domain $\mathcal{D}_i$ and the target domain $\mathcal{D}^{\prime}_j$, respectively.

% \begin{figure}[h]
%   \centering
%   \includegraphics[width=\linewidth]{pics/tvcvscombined.pdf}
%   \caption{The relationship between performance (y-axis) and the combined similarity (x-axis) among source domains and a target domain. Experiments on different target domains are marked with different colors. We also plot the linear relations with lines on the figure for a better intuitive perception of our method. Higher combined similarity leads to better performance. Although there are still a few exceptional cases, it is in general more stable than a single similarity function.}
%   \label{combine}
% \end{figure}

Each similarity function only focus on a single aspect, i.e. the corpus level information or the label level. Therefore, it is inevitable to introduce bias when we select data with them. Naturally, we come up with a strategy that combines all three similarity scores as the indicator to give a more stable guidance for data selection. Assume that one of the combinations, i.e. $C_{\theta_{1},\theta_{2},\theta_{3}}(\text{TVC}_{i}, \text{TIS}_{i}, \text{LO}_{i}) = \theta_{1}\text{TVC}_{i}+\theta_{2}\text{TIS}_{i}+\theta_{3}\text{LO}_{i}$, is linear with the performance, our goal is to find the best value of $\theta_{1}$, $\theta_{2}$, and $\theta_{3}$. For a better reading experience, $C_{\theta_{1},\theta_{2},\theta_{3}}(\text{TVC}_{i}, \text{TIS}_{i}, \text{LO}_{i})$ is abbreviated to $C_{i}$. Following the least squares method  \cite{merriman1877list}, we design the objective function as follows:
\begin{equation}
\begin{array}{l}
    \mathop{\arg\min}\limits_{\theta_1,\theta_2,\theta_3, w, b}  \frac{1}{N_E}\sum_{i=1}^{N_E}\left\|\left[wC_{i} + b \right] - \hat{p}_{i}\right\|^2\\
    \text{s.t.}\quad  w>0, b\geq0 
\end{array}
\label{lstsq}
\end{equation}
where $w$ and $b$ are respectively the weight and bias of the linear function to simulate the linear relation between the indicator score and the performance. $N_E$ is the number of the experiments and $\hat{p}_{i}$ is the true performance of the experiment $i$. $\text{TVC}_{i}$, $\text{TIS}_{i}$, and $\text{LO}_{i}$ are the TVC score, TIS score, and LO score between the source domains and the target domain in the experiment $i$. 

To solve the problem in equation (\ref{lstsq}), we design a scheme to generate samples with the combination of source domains. In general, we pre-define the number of source domains and enumerate all combinations. The three similarity scores between the combination of source domains and target domain will be calculated and recorded. Then we train the model with the combination and record the final performance on the target domain. In this way, we get sufficient tuples $(\verb|TVC|, \verb|TIS|, \verb|LO|, p)$ to figure out the optimum $\theta_1$, $\theta_2$, and $\theta_3$ (see Algorithm~\ref{algorithm}).

% \begin{algorithm}[ht]
% \caption{Training with combination of source domains}\label{algorithm}
% \KwIn{set of source domains $\{\mathcal{D}_1, \cdots , \mathcal{D}_M\}$, target domain $\mathcal{D}^{\prime}$, model $\mathcal{F}$ }
% \KwOut{TVC, TIS, LO, performance $\hat{p}_{i}$}
% \For{$i \leftarrow 1$ \KwTo $M$}{
%     \tcc{all combinations that select $i$ domains from $M$ for training.}
%     $\text{all\_combination} = combination(\{\mathcal{D}_1, \cdots , \mathcal{D}_M\}, i)$\;
%     \For{$j \leftarrow 0$ \KwTo $|{\rm all\_combination}|-1$}{
%         ${\rm combination} = {\rm all\_combination}[j]$\;
%         \tcp{e.g. ${\rm combination} = [\mathcal{D}_1,\mathcal{D}_3]$}
%         $\mathcal{D}_{\text{training}} \leftarrow Merge({\rm combination})$\;
%         $\text{TVC} = TVC(\mathcal{D}_{\text{training}},\mathcal{D}^{\prime})$\;
%         $\text{TIS} = TIS(\mathcal{D}_{\text{training}},\mathcal{D}^{\prime})$\;
%         $\text{LO} = LO(\mathcal{D}_{\text{training}},\mathcal{D}^{\prime})$\;
%         $\text{train}\left(\mathcal{F}(\mathcal{D}_{\text{training}})\right)\text{until Loss converge}$\;
%         $\hat{p}_{i} = \text{eval}(\left(\mathcal{F}(\mathcal{D}^{\prime})\right)$\;
%     }
% }
% \end{algorithm}

\begin{algorithm}[t]
\caption{Training with combination of source domains}
\label{algorithm}
\begin{algorithmic}[1] \footnotesize
    \REQUIRE Set of source domains $\{\mathcal{D}_1, \cdots , \mathcal{D}_M\}$;
    Target domain $\mathcal{D}^{\prime}$;
    Model $\mathcal{F}$;
    \FOR{$1 \le i \le  M$}
    \STATE  $\text{all\_combination} = combination(\{\mathcal{D}_1, \cdots , \mathcal{D}_M\}, i)$\\
    // Select $i$ domain(s) from $M$ for training.
    \FOR{$1 \le j \le  |\text{all\_combination}|-1$}
    \STATE ${\rm combination} = {\rm all\_combination}[j]$\;\\
    // {e.g. ${\rm combination} = [\mathcal{D}_1,\mathcal{D}_3]$}
    \STATE $\mathcal{D}_{\text{training}} \leftarrow Merge({\rm combination})$\;
    \STATE $\text{TVC} = TVC(\mathcal{D}_{\text{training}},\mathcal{D}^{\prime})$\;
    \STATE $\text{TIS} = TIS(\mathcal{D}_{\text{training}},\mathcal{D}^{\prime})$\;
    \STATE $\text{LO} = LO(\mathcal{D}_{\text{training}},\mathcal{D}^{\prime})$\;
    \STATE $\text{train}\left(\mathcal{F}(\mathcal{D}_{\text{training}})\right)\text{until Loss converge}$\;
    \STATE $\hat{p}_{i} = \text{eval}(\left(\mathcal{F}(\mathcal{D}^{\prime})\right)$\;
    \ENDFOR
    \ENDFOR
\end{algorithmic}
\end{algorithm}

With sufficient samples, we fit them with the linear function in equation (~\ref{lstsq}) and optimize $w$, $b$, $\theta_1$, $\theta_2$, and $\theta_3$ via \verb|SGD| \cite{curry1944method}. Due to the data distribution bias of different domains, we finally assign different $w_j$ and $b_j$ for each target domain $\mathcal{D}^{\prime}_j$ to acquire a better linear relation. For the combination weights $\theta_1$, $\theta_2$, and $\theta_3$, we keep same for different target domains. Further, we still have the following points to declare:
\begin{itemize}
\item The parameters $w$ and $b$ are learnable but not necessary for data selection. They are not a part of the indicator and are only used to observe the linear relation between the combination similarity scores and the corresponding performance.
\item Due to the cross-validation setting in the real dataset (e.g. \verb|SNIPS|), to avoid data leakage of the target domain, we obtain $\theta_1$, $\theta_2$, and $\theta_3$ according to the validation domain for each target. The combination form the validation domain still works well on the target and can prove the generality of this strategy.
\item Although training with combination of source domains is time consuming but once the optimum combination weights have been found, it can be adapted to different domains.
\end{itemize}

After that, we can select domains according to the optimum $w^*$, $b^*$, $\theta_1^*$, $\theta_2^*$, and $\theta_3^*$. The domains which can achieve a higher combined similarity score may lead to a better performance and this can be formulated as:
\begin{equation}
\mathop{\arg\max}\limits_{i}\quad w^*\left(\theta_{1}^*\text{TVC}_{i}+\theta_{2}^*\text{TIS}_{i}+\theta_{3}^*\text{LO}_{i}\right) + b^*
\label{selector}
\end{equation}
And due to $w>0$, equation (~\ref{selector}) is equivalent to:
\begin{equation}
\mathop{\arg\max}\limits_{i}\quad\theta_{1}^*\text{TVC}_{i}+\theta_{2}^*\text{TIS}_{i}+\theta_{3}^*\text{LO}_{i}
\label{selector2}
\end{equation}
In this way, the domain specific $w$ and $b$ are eliminated.

\section{Shared-Private Network}
\label{sec_spnet}
\begin{figure*}[h]
  \centering
  \includegraphics[width=\linewidth]{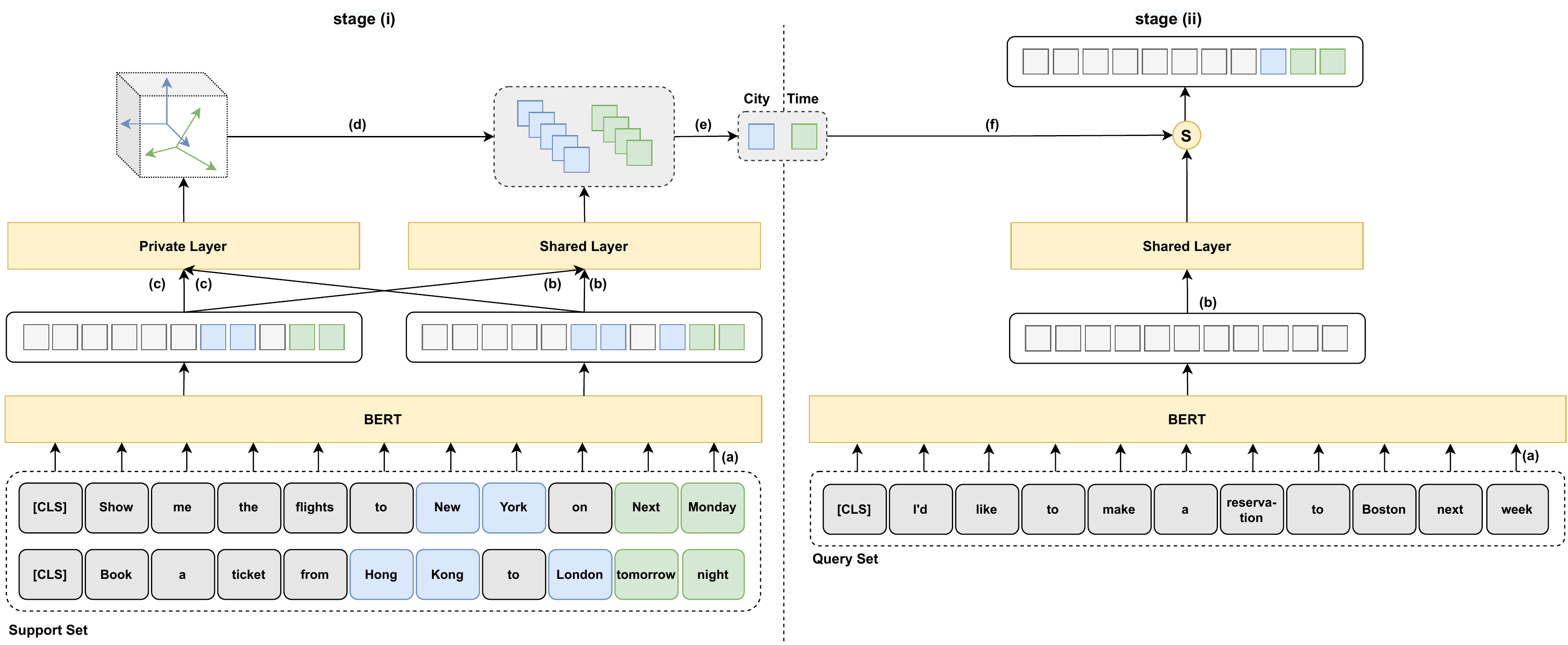}
  \caption{This is the workflow of SP-Net. In this case, the support set contains 2 sentences, and the query set contains 1. The details of processes (a) encode, (b) extract shared features, (c) extract private features, (d) orthogonality constrain, (e) extract label embeddings, and (f) predict are introduced in the main body.}
  \label{overview}
\end{figure*}

Based on the Prototypical Network \cite{snell2017prototypical}, we propose the Shared-Private Network (SP-Net) to gain more representative label embeddings. The workflow is divided into two stages. In the first stage, SP-Net extracts label embeddings for each class from the support set. In the second stage, SP-Net makes prediction on each query sentence according to the label embeddings extracted from stage one. The Figure~\ref{overview} illustrates this process.

\textbf{(a) Encode} Firstly, sentences are encoded into word embeddings via \verb|BERT| \cite{devlin2019bert}. Given a sentence $\textbf{x} = (x_1, x_2, \cdots, x_n)$ as a sequence of words, \verb|BERT| will generate their corresponding contextual word embeddings $\textbf{E} = (E_1, E_2, \cdots, E_n)$, where $E_i \in \mathbb{R}^h$. $h$ is the hidden size of the word embedings.

\textbf{(b) Extract shared features} Although words are different, there are common information among words from the same class. Intuitively, the words in the same class always appear in similar context with similar syntax. And in some cases, they can be even replaced with each other without any grammatical mistakes. For example, even though we replace the phrase ``Hong Kong" with ``New York" in Figure~\ref{overview}, the sentence still makes sense. Common information can help us generate scalable label embeddings that can represent most of the words in a class. The shared layer in the framework is designed for this. In this work, we simply implement the shared layer with a residual linear function and the shared feature of a word is calculated as follows:
\begin{equation}
    E^{s}_i = E_{i} + \text{RELU}(E_{i}W_{s} + b_{s})
\end{equation}
where $W_s \in \mathbb{R}^{h \times h}$ and $b_s \in \mathbb{R}^{h}$ are the weight and bias of the shared layer, respectively. \verb|RELU| is the rectified linear unit function \cite{maas2013rectifier}. 

\textbf{(c) Extract private features} Besides the shared information, each word still has it own specific information. Recall the phrase replacing case mentioned in Figure~\ref{overview}, although the sentence is without any grammatical mistakes after phrase replacing, the meaning has been changed. This is due to the private information carried by the word. The private information is ineffective and can be harmful to label embeddings as they lack generality. Less private information can lead to better quality of label embeddings and therefore, private layer is design to extract private information from the word embeddings. The private layer is also implemented with a residual linear function and the private feature of a word is calculated as fellows:
\begin{equation}
    E^{p}_i = E_{i} + \text{RELU}(E_{i}W_{p} + b_{p})
\end{equation}
where $W_p \in \mathbb{R}^{h \times h}$ and $b_p \in \mathbb{R}^{h}$ is the weight and bias of the private layer, respectively. So far, the shared layer and private layer are symmetrical and share the same design. 

\textbf{(d) orthogonality constrain} To ensure the shared features and private features are separated completely, we introduce the following constrains:
\begin{itemize}
\item The shared features of the words in a same class should be close to each other.
\item The private features of words should be diverse even though they belong to the same class.
\item The shared feature and the private feature of a word should not overlap.
\end{itemize}

For the first requirement, \citet{chen2020simple} proposed to use contrastive loss that can make the same samples to be close and different samples to be far apart. The similarity between samples are defined as:
\begin{equation}
    \text{sim}(E_i^{s}, E_j^{s}) = \frac{E_i^{s\top}E_j^{s}}{\|E_i^{s}\|\|E_j^{s}\|}
\end{equation}
The loss in the first requirement is defined as:
\begin{equation}
    \mathcal{L}_1 =\mathop{\mathbb{E}}\limits_{c}\left[ -\log\frac{\sum_{\{i;y_i=c\}}\sum_{\{j;y_j=c\}}{\exp({\text{sim}(E_i^{s},E_j^{s})}/\tau)}}
    {\sum_{\{i;i\in \mathcal{S}\}}\sum_{\{j;j\in \mathcal{S}\}}{\exp({\text{sim}(E_i^{s},E_j^{s})}/\tau)}}\right]
\end{equation}
where $\tau$ is the temperature parameter and $c$ is the class. The numerator is the sum of the similarity scores whose class is $c$. The denominator is the sum of all the similarity scores. Specifically, embeddings in the same class presents high similarity score and the numerator is large and the loss decreases.

\begin{table*}[ht]
\begin{adjustbox}{max width=\textwidth}
\begin{tabular}{l|lllllllll}
\hline
 & \textbf{Model} & \textbf{We} & \textbf{Mu} & \textbf{Pl} & \textbf{Bo} & \textbf{Se} & \textbf{Re} & \textbf{Cr} & \textbf{Avg.} \\ \hline
\multicolumn{1}{c|}{\multirow{0}{*}{\textbf{1-shot}}} & SimBERT & 36.10 & 37.08 & 35.11 & 68.09 & 41.61 & 42.82 & 23.91 & 40.67 \\
\multicolumn{1}{c|}{} & TransferBERT & 55.82 & 38.01 & 45.65 & 31.63 & 21.96 & 41.79 & 38.53 & 39.06 \\
\multicolumn{1}{c|}{} & L-TapNet+CDT+PWE \cite{hou2020fewshot} & 71.53 & 60.56 & 66.27 & \textbf{84.54} & 76.27 & 70.79 & 62.89 & 70.41 \\
\multicolumn{1}{c|}{} & L-ProtoNet+CDT+VPB \cite{zhu2020vector} & 73.12 & 57.86 & 69.01 & 82.49 & 75.11 & 73.34 & 70.46 & 71.63 \\
\multicolumn{1}{c|}{} & SP-Origin & 70.67 & 59.27 & 69.58 & 82.80 & 76.92 & 72.49 & 74.63 & 72.34 \\
\multicolumn{1}{c|}{} & SP-Domain Selection & \textbf{76.07} & \textbf{64.29} & \textbf{71.10} & 84.19 & \textbf{81.63} & \textbf{73.66} & \textbf{76.41} & \textbf{75.34} \textcolor{mypink1}{(+3.00)} \\ \hline
\multirow{0}{*}{\textbf{5-shot}} & SimBERT & 53.46 & 54.13 & 42.81 & 75.54 & 57.10 & 55.30 & 32.38 & 52.96 \\
 & TransferBERT & 59.41 & 42.00 & 46.07 & 20.74 & 28.20 & 67.75 & 58.61 & 46.11 \\
 & L-TapNet+CDT+PWE\cite{hou2020fewshot} & 71.64 & 67.16 & 75.88 & 84.38 & 82.58 & 70.05 & 73.41 & 75.01 \\
 & L-ProtoNet+CDT+VPB\cite{zhu2020vector} & 82.93 & 69.62 & 80.86 & \textbf{91.19} & 86.58 & \textbf{81.97} & 76.02 & 81.31 \\
 & SP-Origin & 83.92 & 69.37 & 79.47 & 89.43 & 87.95 & 77.75 & 80.31 & 81.17 \\
 & SP-Domain Selection & \textbf{84.03} & \textbf{71.09} & \textbf{82.01} & 90.13 & \textbf{89.44} & 80.71 & \textbf{80.88} & \textbf{82.61} \textcolor{mypink1}{(+1.44)} \\ \hline
\end{tabular}
\end{adjustbox}
\caption{F1 scores of few-shot slot tagging on SNIPS dataset}
\label{result}
\end{table*}

\begin{table*}[]\centering
\begin{adjustbox}{max width=\textwidth}
\begin{tabular}{lllllllllll}
\hline
\multirow{2}{*}{\textbf{Model}} & \multicolumn{5}{c}{\textbf{1-shot}}                                                                     & \multicolumn{5}{c}{\textbf{5-shot}}                                                 \\ \cline{2-11} 
                                & \textbf{News}  & \textbf{Wiki}  & \textbf{Social} & \multicolumn{1}{r}{\textbf{Mixed}} & \textbf{Avg.}  & \textbf{News}  & \textbf{Wiki}  & \textbf{Social} & \textbf{Mixed} & \textbf{Avg.}  \\ \hline
SimBERT                         & 19.22          & 6.91           & 5.18            & \multicolumn{1}{r}{13.99}          & 11.32          & 32.01          & 10.63          & 8.20            & 21.14          & 18.00          \\
TransferBERT                    & 4.75           & 0.57           & 2.71            & \multicolumn{1}{r}{3.46}           & 2.87           & 15.36          & 3.62           & 11.08           & 35.49          & 16.39          \\
L-TapNet+CDT+PWE                & 44.30          & \textbf{12.04} & 20.80           & 15.17                              & 23.08          & 45.35          & 11.65          & 23.30           & 20.95          & 25.31          \\
L-ProtoNet+CDT+VPB              & 43.47          & 10.95          & \textbf{28.43}  & \textbf{33.14}                     & 29.00          & 56.30          & 18.57          & \textbf{35.42}  & \textbf{44.71} & 38.75          \\ \hline
SP-Origin                       & \textbf{43.50} & 12.02          & 27.77           & 33.05                              & \textbf{29.08} & \textbf{57.70} & 18.62          & 35.41           & 44.67          & 39.10          \\
SP-Domain Selection             & 43.50          & 12.02          & 27.77           & 33.05                              & 29.08 \textcolor{mypink1}{(+0.00)}          & 57.70          & \textbf{21.11} & 35.41           & 44.67          & \textbf{39.72} \textcolor{mypink1}{(+0.62)}\\ \hline
\end{tabular}
\end{adjustbox}
\caption{F1 scores of few-shot slot tagging on NER dataset}
\label{result2}
\end{table*}

For the second requirement, according to the co-variance of two variables, we define the divergence between two embeddings as:
\begin{equation}
    D(E_i^{p}, E_j^{p}) = (E_i^{p} - \mathbb{E}^p)^T(E_j^{p} - \mathbb{E}^p)
\end{equation}
where $\mathbb{E}^p$ is the mean vector of all private embeddings in the set. The loss in the second requirement is:
\begin{equation}
\label{l2}
    \mathcal{L}_2 = -\frac{1}{|\mathcal{S}|^2}\sum_{i\in \mathcal{S}}{\sum_{j\in \mathcal{S}}{\log{D(E_i^{s}, E_j^{s})}}}
\end{equation}
where $|\mathcal{S}|$ is the size of the support set, i.e. the number of words. Higher divergence among the private embeddings will lead to lower loss. We also implement \verb|L2-norm| to restrain the increase of the parameters.

The third requirement refines the shared features further. We introduce the orthogonality constraints \cite{liu-etal-2017-adversarial} to force the shared embedding independent with the private embedding:
\begin{equation}
\label{l3}
    \mathcal{L}_3 = \frac{1}{|\mathcal{S}|}\sum_{i\in \mathcal{S}}{\left\|E_i^{s\top}E_i^{p}\right\|_2}
\end{equation}
where $\|\cdot\|_2$ is the Euclidean norm. 

\textbf{(e) Extract label embeddings} Label embeddings are extracted from shared embeddings for each class. We take the mean vector of the shared embeddings which belong to class $c$ as the label embedding:
\begin{equation}
    E^c = \frac{1}{|\{y_i=c\}|}\sum_{\{y_i=c\}}{E^s_i}
\end{equation}
where $E^c$ is the label embedding of the class $c$. 

\textbf{(f) Predict} We calculate the similarity between shared embeddings of the query sentence with the label embeddings. We provide various options and here we take cosine similarity as an example:

\begin{equation}
\label{cosine}
    p^c_i = \frac{E^{s\top}_{i}E^c}{\|E^{s}\|\|E^{c}\|}
\end{equation}

where $p^c_i$ is the similarity between word $i$ with class $c$ and can also be regarded the confidence that the word belongs to this class. The class with the highest similarity will be regarded as the prediction for the word. We take the binary cross-entropy loss to measure the error in each class:
\begin{equation}
    \mathcal{L}_4 = \frac{1}{|\mathcal{Q}|}\sum_{i}^{|\mathcal{Q}|}\sum_{c}^{C}y_i\log{p^c_i} + (1-y_i)\log{(1-p^c_i)}
\end{equation}
where $C$ is the number of unique labels in the query set and $|\mathcal{Q}|$ is the number of words in the query set.

Finally, we combine the $\mathcal{L}_1$, $\mathcal{L}_2$, $\mathcal{L}_3$, and $\mathcal{L}_4$ with different weights as the cost function:

\begin{equation}
\label{loss}
    \mathcal{L} = \alpha\mathcal{L}_1 + \beta\mathcal{L}_2 + \gamma\mathcal{L}_3 + \delta\mathcal{L}_4
\end{equation}

where $\alpha$, $\beta$, $\gamma$, and $\delta$ are hyperparameters determined by the experiments.

\section{Experiments}
\label{sec_exp}

\subsection{Dataset}
We evaluate the proposed method following the same experiment setting provided by \citet{hou2020fewshot} on SNIPS \cite{coucke2018snips} and NER dataset \cite{zhu2020vector}. SNIPS contains 7 domains including Weather (We), Music (Mu), PlayList (Pl), Book (Bo), Search Screen (Se), Restaurant (Re), and Creative Work (Cr) and the sentences in SNIPS are annotated with token-level \verb|BIO| labels for slot tagging. Each domain will be tested in turn following cross-validation strategy. In each turn, 5 domains are used for training and 1 for evaluation. In each domain, the data are split into 100 episodes \cite{ren2018meta}. For the sake of fair peer comparison, the selection of evaluation domain and episodes construct are kept same with \citet{hou2020fewshot}. NER dataset contains 4 domains including News, Wiki, Social, and Mixed. In addition, due to the number of domains in the NER dataset is too short, we randomly split domains into pieces and select those pieces via the combined similarity function. More training details can be found in the appendix.

\subsection{Baselines}
\noindent \textbf{SimBERT} assigns label to the word according to cosine similarity of word embedding of a fixed BERT. For each word $x_i$ , SimBERT ﬁnds the most similar word $x_k$ in the support set and assign the label of $x_k$ to $x_i$.

\noindent \textbf{TransferBERT} directly transfers the knowledge from source domain to target domain by parameter sharing. 

\noindent \textbf{L-TapNet+CDT+PWE} \cite{hou2020fewshot} is a strong baseline for few-shot slot tagging that combines with the label name representation and a special CRF framework.

\noindent \textbf{L-ProtoNet+CDT+VPB} \cite{zhu2020vector} investigates different distance functions and utilizes the powerful distance function VPB to boost the performance of the model. 

\noindent \textbf{SP-Net } is proposed in this work that utilizes the Shared-Private layer to capture the common features and generate a more stable label representation.

\noindent \textbf{SP-Net + Domain Selection } is also SP-Net but it is trained with the selected data according to the data selection strategy we proposed.

\subsection{Main Results}

Table \ref{result} shows the results of 1-shot and 5-shot on the SNIPS dataset. Generally speaking, the SP-Net achieves best performance on the 1-shot setting and comparable performance on the 5-shot setting (0.14\% adrift of SOTA). As for the data selection strategy, it greatly enhances the performance on both of the 1-shot and 5-shot settings. With the data selection, the performance of SP-Net is far beyond other baselines.

The result on the NER dataset also prove the effectiveness of our method (See Table \ref{result2}). It is noticed that, due to the short of the data, combined similarity select all data on most domains except Wiki of 5-shot task. Therefore the result of SP-Origin and SP-Domain Selection are nearly the same.

\begin{figure}[h]
  \centering
  \includegraphics[scale=0.3]{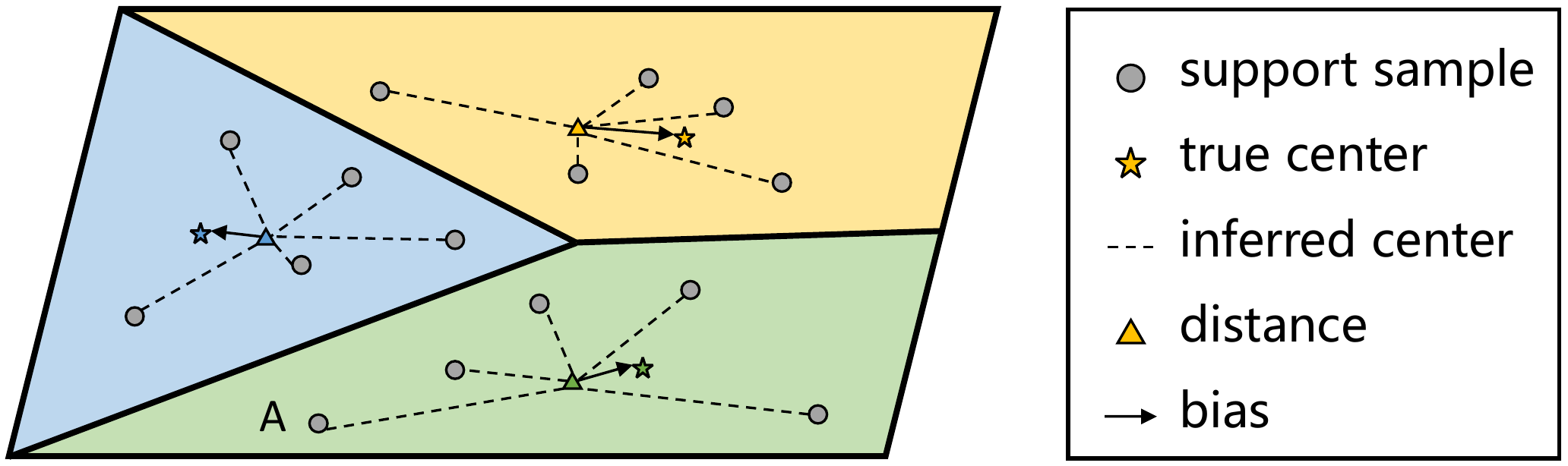}
  \caption{This is diagram shows the automatic correction of distribution bias when the number of supports increased. The circles are samples in the support set and triangles are the inferred center, as well as label embedding, according to the supports. Stars are the true center of classes.}
  \label{bias}
\end{figure}

The effect of Shared-Private Network is more remarkable if the number of the support samples is less. The SP-Net outperforms all baseline in the 1-shot setting but in 5-shot, it achieves comparable performance. The shared-private Network, essentially, corrects the bias between the label embedding and the center of the class. The bias will be more serious if the support is less. With the increase of the number of supports, bias could be suppressed to some extent (see Figure~\ref{bias}). Some other methods, like label description \cite{hou2020fewshot}, can also correct such kind of bias if enough supports are given. But when the supports are extremely scarce, Shared-Private Network performs the best.

% The effect of data selection is also more effective on the 1-shot setting than 5-shot. Data selection is a noise reduction process before training. When the supports are less, the interference from noise is more serious. Therefore, data selection improves the performance more on 1-shot. Another potential reason is that we select domains rather than episodes or samples. Each domain contains 100 episodes and if one domain is regarded ineffective and not used in the training, the volume of training data will be decreased, which may weaken performance. The loss of the training data is higher for 5-shot.

\subsection{Analysis}

\begin{figure}[h]
  \centering
  \includegraphics[width=\linewidth]{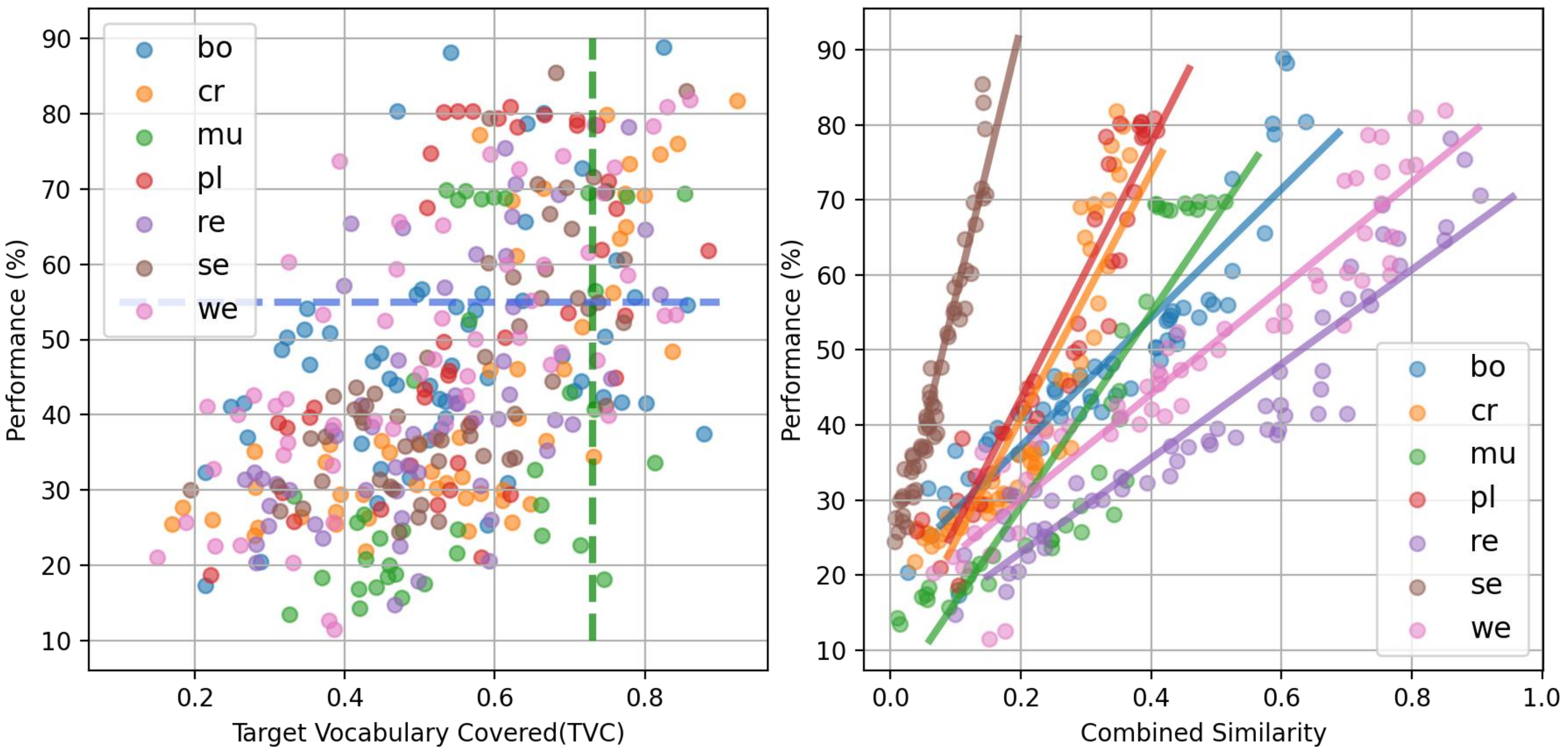}
  \caption{The relation between performance (y-axis) and the similarity function (y-axis). Different target domains are in different colors.}
  \label{tvcvscombined}
\end{figure}

We further visualize the relation between the performance with the similarity function and compare combined similarity with \verb|TVC| in Figure~\ref{tvcvscombined}. We firstly sample some combinations of source domains and train the model. Then we calculate their similarity with the target domain and record performance. From the left part of Figure~\ref{tvcvscombined}, the performance generally has a positive correlation with \verb|TVC|. However, its precision is poor so that cannot be used as an indicator. Points around the green line have similar \verb|TVC| scores but the performance are quite diverse, i.e. the performance of green points' are from 20\% to 70\%. A similar conclusion can be drawn from the horizontal direction: blue points around the blue line have similar performance but their \verb|TVC| scores are from 36\% to 87\%. Therefore, data selection with \verb|TVC| suffers from serious performance fluctuation. By comparison, there is an apparent positive linear correlation between combined similarity and performance in terms of target domain (See the right part of Figure~\ref{tvcvscombined}). 

\begin{figure}[h]
  \centering
  \includegraphics[width=\linewidth]{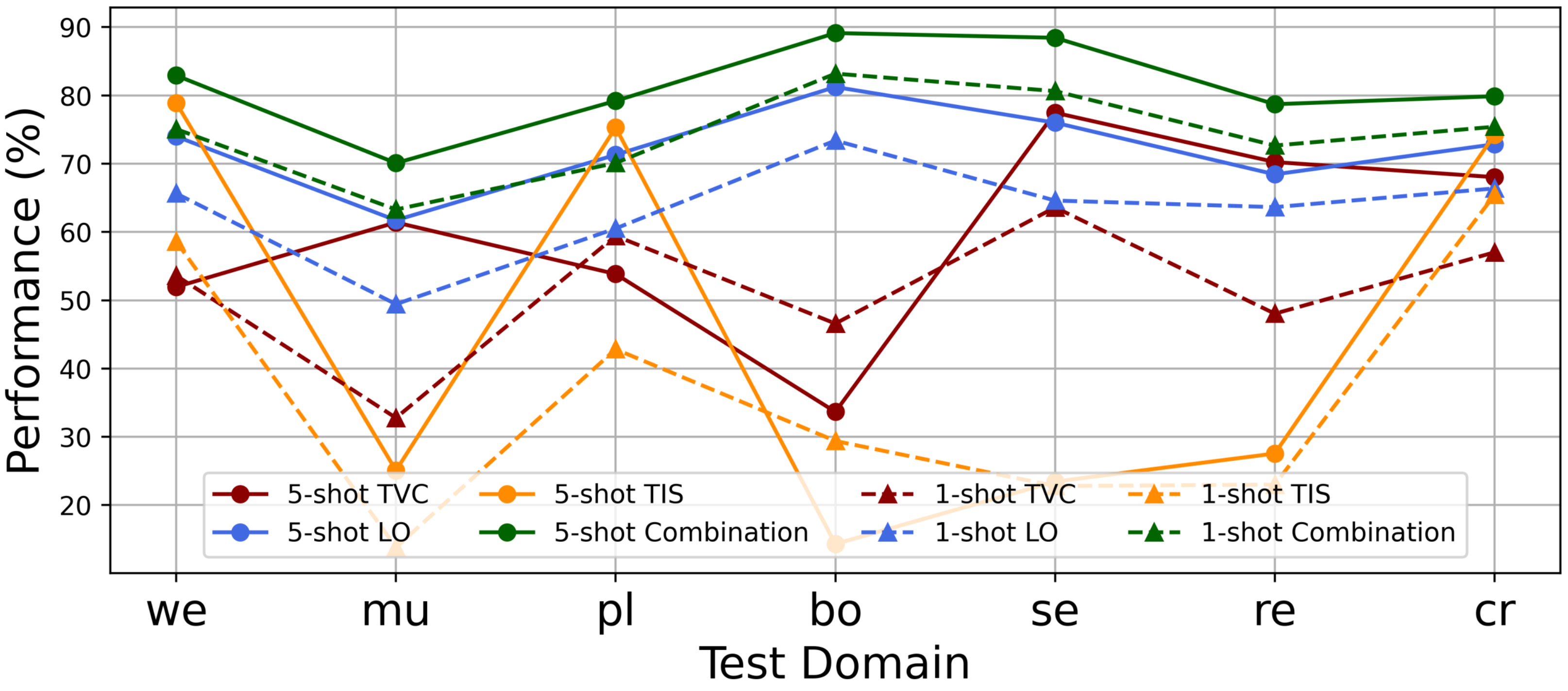}
  \caption{The performance of training with domains selected by 4 functions.}
  \label{sim_compare}
\end{figure}

In order to prove the advantage of the combination similarity function, we compare it with its component \verb|TVC|, \verb|TIS|, and \verb|LO|. The result is shown in Figure~\ref{sim_compare}. The performance of our combination similarity function (the green line) outperforms others on both 1-shot and 5-shot. Besides that, the \verb|LO| similarity (blue line) performs equally on different test domain, which is more stable than \verb|TVC| and \verb|TIS|. By contrast, the performance of \verb|TVC| and \verb|TIS| have huge variance on various test domains. Sometimes they can surpass \verb|LO| and sometimes their performance even lower than 20\%. This is because the 3 similarity functions have their own pros and cons and the combination of them is more effective and stable (See Appendix for more analysis about iter-domain relation). 

\section{Conclusions and Future Work}
\label{sec_cf}
In this paper, we prove the existence of negative knowledge transfer in few-shot learning and propose a similarity-based method to select proper data before training. We propose a Shared-Private Network (SP-Net) for the few-shot slot tagging task. We prove the effectiveness and advantages of both data selection method and SP-Net with experiments. In the future, we will investigate the relations among domains and improve our data selection method to select episodes or samples rather than domains. Also, we will analysis and explain SP-Net from the latent space to figure out what it exactly correct for the label embeddings.

\bibliography{aaai22.bib}
\end{document}

\appendix
\section{Appendix}
\subsection{Inter-domain relations} 

\begin{figure}[h]
  \centering
  \includegraphics[width=\linewidth]{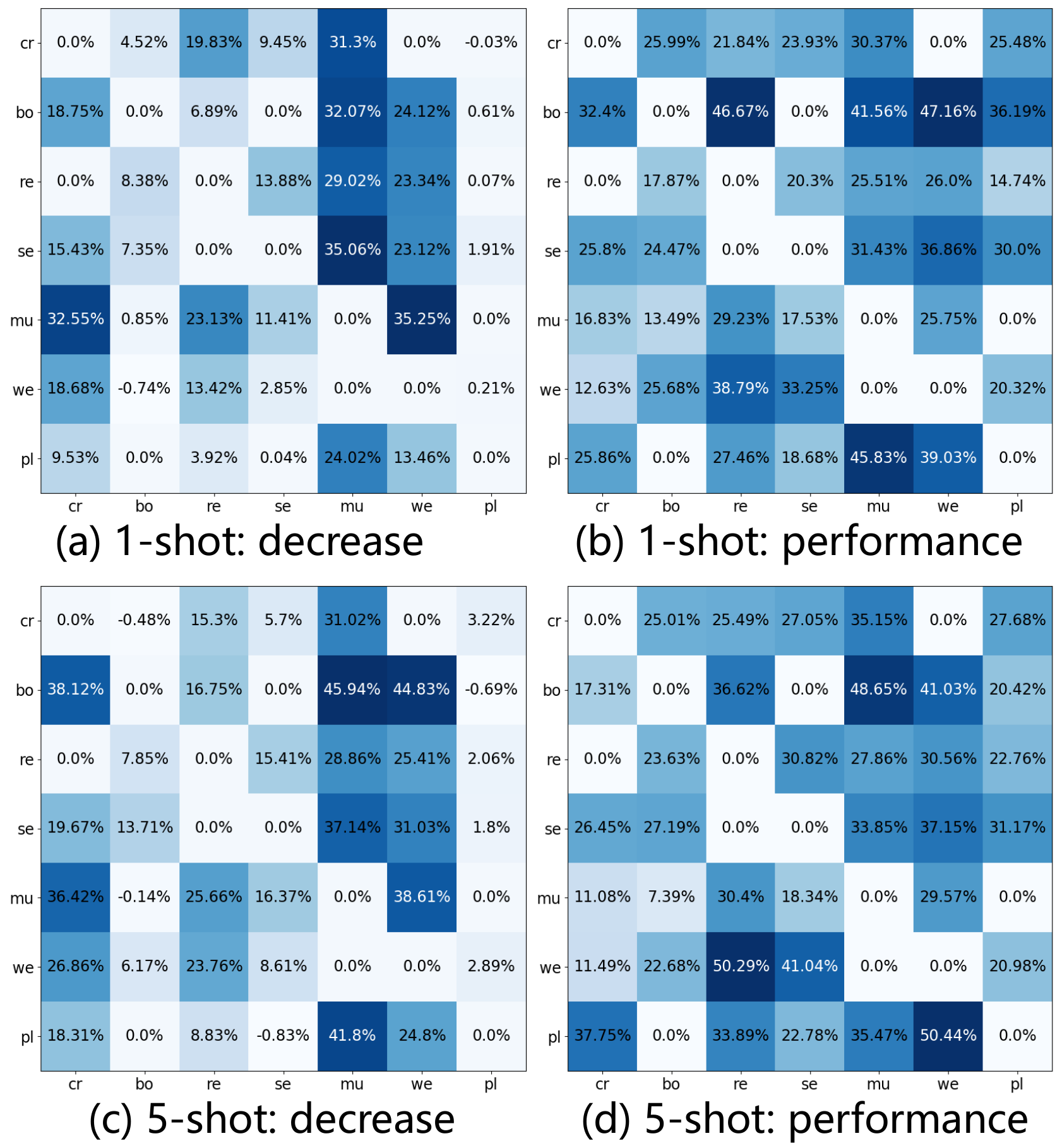}
  \caption{The heat map shows the inter-domain relations. The y-axis is the target domain and the x-axis is the source domain. The picture (a) and (b) are the results of 1-shot setting. The picture (c) and (d) are the results of 5-shot setting. The picture (a) and (c) illustrate the performances' decreases on the target if a source domain is removed. The (b) and (d) illustrate the performance on the target, which the model is trained with a single source domain.}
  \label{heatmap}
\end{figure}

We further study the inter-domain relations which can give strong evidences to prove the importance of data selection. We have a key assumptions in this part: If a source domain and a target domain have a strong relation, (1) removing the source domain from training will decrease the performance on the target domain or (2) training with the single source domain will have a better performance than training with a unrelated domain. Following these two assumptions, we conduct two experiments: (1) For every test, remove each domain from the 5 training domains in turn, train SP-Net, and then record the performance; (2) For every test, select each domain from the 5 training domains in turn, train SP-Net, and then record the performance. Figure ~\ref{heatmap} shows the results and we have two findings.

Firstly, the differences of the source domains have a significant influence to the final performance. For example, in Figure ~\ref{heatmap} (a), if the source domain \verb|mu| is removed from training, 35.06\% performance decreased is observed in target (test) domain \verb|se|. By comparison, for the same test domain \verb|se|, the removal of domain \verb|pl| causes 1.91\% decrease, which is slighter. Similarly, in Figure ~\ref{heatmap} (d), only with the domain \verb|re|, the performance on target domain \verb|we| can achieve 50.29\%. By contrast, with the domain \verb|cr|, the performance on \verb|we| only has 11.49\%. Different source domains bring huge variance in performance. This result shows the need of data selection. Secondly, some negative values appeared in Figure ~\ref{heatmap} (a) and (c), which means after removing a domain, the performance is improved. For instance, in Figure ~\ref{heatmap} (c), removing the domain \verb|se| leads to 0.83\% increase (-0.83\% decreases). This phenomena gives another strong evidence of negative knowledge transfer.

\subsection{Training Details}

\textbf{Hyperparameters} The \verb|BERT| in SP-Net is the pre-trained uncased \verb|BERT-Base| \cite{devlin2019bert}. We use \verb|ADAM| \cite{kingma2014adam} to train the model with a learning rate of 2e-5, a weight decay of 5e-5. And we set \verb|VPB| \cite{zhu2020vector} as the similarity function for prediction. For the weights assigned to each loss, we set $\alpha$, $\beta$, $\gamma$, and $\delta$ as 0.2, 0.1, 0.2, and 0.5 respectively. Those hyperparameters mentioned above are derived from the best implement in our experiments. To prevent the impact of randomness, we do each experiment 10 times with different random seed and report the average results. 

\noindent\textbf{Data Selection} Due to cross validation, each domain is used in turn as a test domain. As such one domain used for training and may be used for testing next time. Therefore, if we set a group of global similarity combination weights $\theta_1$, $\theta_2$, and $\theta_3$ according to all experimental results, it must lead to test data leakage. This is unfair for the comparison. To this end, we set $\theta_1$, $\theta_2$, and $\theta_3$ in terms of the test domain, respectively. $\theta_1$, $\theta_2$, and $\theta_3$ is obtained by minimizing Equation (8) according to the training domains and evaluation domain. In addition, if $\theta_1$, $\theta_2$, and $\theta_3$ from the evaluation domain work well in the test domain, it demonstrates the generality of this data selection method. In practice, the combination weights just need to be calculated once. In this work, we set a domain as the minimum selection unit. Specifically, if a domain is selected for training, all episode in this domain will be selected. The domain selection follows Equation (10).

\bibliography{aaai22.bib}